\documentclass[aps, twocolumn]{revtex4}
\usepackage{graphicx}
\usepackage{tikz}

\begin{document}

\title{Monte Carlo Sort for unreliable human comparisons}
\author{Samuel L Smith}
\email{samuel.laurence.smith@gmail.com}

\affiliation{Theory of Condensed Matter, Cavendish Laboratory, University of Cambridge}
\affiliation{babylon health, 60 Sloane Avenue, London, SW3 3DD}
\affiliation{ASI Data Science, 48 Fitzroy Street, London, W1T 5BS}

\begin{abstract}
Algorithms which sort lists of real numbers into ascending order have been studied for decades. They are typically based on a series of pairwise comparisons and run entirely 'on chip'. However people routinely sort lists which depend on subjective or complex judgements that cannot be automated. Examples include marketing research; where surveys are used to learn about customer preferences for products, the recruiting process; where interviewers attempt to rank potential employees, and sporting tournaments; where we infer team rankings from a series of one-on-one matches. We develop a novel sorting algorithm, where each pairwise comparison reflects a subjective human judgement about which element is 'bigger' or 'better'. We introduce a finite and large error rate to each judgement, and we take the 'cost' of each comparison to significantly exceed the cost of other computational steps. The algorithm must request the most informative sequence of comparisons from the user; in order to identify the correct sorted list with minimum human input. Our "Discrete Adiabatic" Monte Carlo approach exploits the gradual acquisition of information by tracking a set of plausible hypotheses which are updated after each additional comparison.
\end{abstract}

\maketitle

\section{Introduction}
Every undergraduate in computer science is familiar with many algorithms for sorting lists. All these algorithms are based on a series of pairwise comparisons between two elements, and they form the backbone of many important applications. Two assumptions underlie most approaches. First, we assume that the computer does not make mistakes; if the computer measures that $a<b$, then $a<b$ \cite{feige1990computing, busse2012information, CPC:248097, ajtai2009sorting, hadjicostas2011recursive, yao1985fault}. Second, we assume that the cost of comparing two elements is not significantly greater than the cost of any other computational operation; our goal is to minimise the total number of operations required to sort the list, which does not necessarily mean we will minimise the number of pairwise comparisons.

Now consider a marketing company, which wishes to learn about its customers' sporting preferences. It could produce a survey which asks \textit{"on a scale of 1 to 10, how much do you enjoy the following activities?..."}. Such surveys are not only tedious, they are also very difficult for the user to complete, since they ask people to rank their preferences on an abstract scale which does not exist. However what the company is really trying to obtain is an ordered list of user preferences, so they could instead ask a series of pairwise comparisons like \textit{"do you prefer golf or tennis?"}. This question is well defined, and much easier for the user to answer. The goal of this paper is to develop an algorithm which infers ordered lists from pairwise subjective judgements.

To continue, we must specify a probability model to describe the comparisons, and decide what the algorithm will optimise. Since a customer's time is valuable, and computers can perform billions of calculations per second; we wish to minimise the number of human judgements required to find the true sorted list. Since people are prone to error, the simplest possible model of each judgement is the noisy channel, whereby the probability that a user returns a correct answer is $p$, and the probability that a user returns an incorrect answer is $(1-p)$. Initially we will assume that $p$ is known, but we will relax this assumption at the end of the paper.

In this simple model, the probability of an error is independent of the elements being compared. We will consider element-dependent error models in a future paper. Strictly speaking, we can never be sure we have found the correct sorted list, and instead we must terminate the algorithm when the probability we have made an error is sufficiently low. Computer time is not free, so we will ensure that the computational cost of the algorithm is practical for lists of $10$-$1000$ elements. In order to minimise the number of comparisons required, the sequence of questions must be chosen adaptively. We will take a "greedy entropy" approach, and develop an algorithm which at each point in time chooses the comparison with the highest expected information content. Since there are two possible responses to each comparison, the expected information content is maximised when the probability of each response is 1/2.

This simple problem is emblematic of a wide range of challenges. People remain better than computers at solving many complex problems, yet computers are far superior to people at evaluating inference models. Human-computer interfaces have great potential to minimise the well-known human biases which impede decision making.

\section{Bayes theorem and rejection sampling}
Consider sorting just two elements $\{ab\}$. Throughout this paper, we will assume uniform priors, such that the initial probabilities $P_0(ab) = 1/2$ and $P_0(ba) = 1/2$. Imagine we measure $a<b$. The updated probabilities,
\begin{eqnarray}
P_1(ab) &=& p/Z_1, \nonumber \\
P_1(ba) &=& (1-p)/Z_1.
\end{eqnarray} 
The normalisation constant $Z_1 = p + (1-p) = 1$. Now imagine we have made $n$ measurements, $n_{match}$ of which found $a<b$, while the remainder found $b<a$. The probability of the ordering 'ab' is now $P_n(ab) = p^{n_{match}}(1-p)^{n-n_{match}}/Z_n$. This is a simple example of Bayes theorem, whereby
\begin{eqnarray}
P_n(\text{order}|\text{data}) &=& P_n(\text{data}|\text{order})/Z_n \nonumber \\
                                 & = & p^{n_{match}} (1-p)^{n - n_{match}} / Z_n.
\end{eqnarray}
This result applies to lists of arbitrary length, where $n_{match}$ denotes the total number of measurements which have been consistent with the order in question. The normalisation constant $Z_n = \sum_{all\; orders} P_n(\text{data}|\text{order})$. This sum is evaluated over the $L!$ possible orderings of the $L$ elements in the list. If $L$ is small, then we can evaluate $Z_n$ explicitly. When $L$ is large we cannot, and so we cannot calculate $P_n(\text{order}|\text{data})$. However we can draw valid samples from it by rejection sampling. A naive rejection sampler would proceed as follows,
\begin{itemize}
\item Draw a sample order from the uniform distribution.
\item Accept the sample with probability $(\frac{1-p}{p})^{n-n_{match}}$, if the sample is rejected return to step 1.
\end{itemize}
Using this sampler, we can develop a simple Monte Carlo algorithm to sort the list,
\begin{itemize}
\item Draw a set of N candidate orderings from $P_n(\text{order}|\text{data})$.
\item Identify the two elements $\{ij\}$ about which the candidate orderings are most uncertain, such that we minimise $( N_{ij} - N_{ji})^2$, where $N_{ij}$ is the number of candidate lists with $i<j$.
\item Ask the user if $i <j$, set $n \rightarrow n+1$, and return to step 1. Terminate the algorithm when all N candidate orderings are the same list.
\end{itemize}
This algorithm uses the candidate lists both to estimate which question has the highest expected information content, and identify when the algorithm has converged. However the naive rejection sampler is prohibitively computationally expensive. To see this, we note that sample lists are drawn uniformly from the $L!$ possible orderings, while the algorithm will only terminate when all N candidate lists are identical. Consequently we expect to draw at least $O(N \times L!)$ samples to reach convergence. Additionally, it seems irrational to throw away our candidate lists after each measurement, since they might contain some useful information. In the remainder of this section we will propose three more efficient sampling strategies. In the following section we introduce our complete algorithm and evaluate its performance.

\subsection{Sampling by recursion}
Our naive rejection sampler has two related flaws. First, it requires the computer to draw the correct list randomly and uniformly from the $L!$ possible orderings. Second, as the number of measurements $n$ becomes large, the probability that a uniform random sample is accepted will become vanishingly small. In recursion sampling, we do not ask the computer to generate a candidate list in one go, we only ask it to partition the list into two sets of elements. The algorithm proceeds as follows,
\begin{itemize}
\item Randomly assign half of the elements to the set $X_{<}$ and half to the set $X_>$.
\item Count the number of times $n_{dispute}$ that an element in set $X_<$ has been measured to be bigger than an element in set $X_>$.
\item Accept the partition with probability $(\frac{1-p}{p})^{n_{dispute}}$. If the partition is rejected, return to step 1. If the partition is accepted, then elements in $X_<$ will appear below elements in $X_>$ in the final sample.
\item Recursively generate partitions for the two subsets $X_<$ and $X_>$, terminating the recursion when we reach the trivial set containing a single element.
\end{itemize}
This recursive strategy improves on the naive rejection sampler, because we are now able to make incremental progress towards generating a sample; if a sample partition is rejected, we do not start from scratch, but simply try another partition for that subsection of the overall list. The final sample remains a numerically exact sample from $P_n(\text{order}|\text{data})$.

\subsection{Sampling by maximum element}
An alternative strategy also utilises our freedom to generate a sample from a series of partitions. However instead of generating two partitions of similar size, we instead repeatedly sample the largest element from the list. The algorithm proceeds as follows,
\begin{itemize}
\item Select a single random element $e_{max}$ uniformly from the list.
\item Count the number of times $n_{dispute}$ that $e_{max}$ has been measured to be smaller than another element.
\item Accept $e_{max}$ as the largest element in the sample with probability $(\frac{1-p}{p})^{n_{dispute}}$. If $e_{max}$ is rejected, return to step 1.
\item If $e_{max}$ is accepted, then it will be the largest element in the final sample. Remove it from the list, and repeat the above on the new sublist. Continue until we reach the trivial sublist containing a single element.
\end{itemize}
There are now only L possible partitions of the list, and we can reformulate this procedure using the ideas of arithmetic coding. We modify the algorithm as follows,
\begin{itemize}
\item Set a float w = 0, integer i = 1, and draw a random number $0 \leq \sigma \leq 1$. Assume that the elements are stored in memory in some order.
\item Evaluate $\beta = (\frac{1-p}{p})^{n_{dispute}}$ for each of the L possible elements, and compute the normalisation constant $Z_p = \sum_j  \beta_j$.
\item Evaluate $\gamma = \beta_i/Z_p$. If $\sigma < w + \gamma$, accept the element. If not, let $w \rightarrow w + \gamma$, $i \rightarrow i + 1$, and repeat step 3 until an element is accepted.
\end{itemize}
We need only generate one random number in order to select the next largest element. Yet we still generate perfect final samples from $P_n(\text{order}|\text{data})$, so long as the precision of this random number is sufficiently small. It takes $O(L)$ time to choose a maximum element\footnote{Naively it appears to take $O(L^2)$ to choose the maximum element, since we must obtain $n_{dispute}$ for each of $L$ elements, and each element may have been compared with $(L-1)$ other elements. However this is easily resolved. Store a value of $n_{dispute}$ for each element, as well as a list of all the elements that element has "beaten" in a measurement. Once the maximum element has been chosen, use this second list to update $n_{dispute}$ for the remaining elements. Since we expect to require $O(\ln L)$ measurements per element to sort the list, this update step is also $O(\ln L)$, and it only occurs once per partition. The most expensive step remains evaluating the normalisation constant, which is now $O(L)$ since $n_{dispute}$ was already known for each element. The run-time is weakly dependent on the number of measurements that have been made, but only in a subleading term which becomes irrelevant as $L\rightarrow \infty$.}, and will therefore take $O(L^2)$ time to generate a complete sample ordering of the whole list. Unlike the naive sampler, this run-time is essentially independent of the number of measurements $n$ that have been made. Simple Markov Chain Monte Carlo implementations would also exhibit $O(L^2)$ run-times, but we cannot be certain that these will generate a numerically perfect sample.

We can put the elements in memory in whatever order we wish. Ideally the elements will lie roughly in order of increasing $n_{dispute}$. In this case, it will be sensible to sum the normalisation constant backwards through the set in order of increasing likelihood (improving the accuracy of the summation). However we then run the selection step forwards in order of decreasing likelihood, to minimise the number of operations required to select an element.

\subsection{Sampling from previous candidates}
In the simple Monte Carlo approach, after each new measurement arrived, we drew completely new random samples from $P_{n+1} = P_{n+1}(\text{order}|\text{data})$. This is foolish. The previous set of candidates were drawn from $P_n$. Imagine that we just measured $a<b$. Then
\begin{eqnarray}
P_{n+1} \propto  P_n  \times p &&\text{ if $a<b$ in the sample,} \nonumber \\
                      P_{n+1} \propto  P_n \times (1-p) &&\text{ if $b<a$ in the sample.}
\end{eqnarray}
We can accept any candidate drawn from $P_n$ which satisfies $a<b$ as a valid sample from $P_{n+1}$, while also accepting candidates drawn from $P_n$ which satisfy $b<a$ with probability $(1-p)/p$. Naively we must regenerate any candidate that fails this test from scratch, however we have already seen that samples can be generated from a series of partitions. Consider an example candidate, drawn from $P_n$, with ordering $\{q...wb...ar...t\}$. This list can be split into three partitions; $\{q...w\}$, $\{b...a\}$, and $\{r...t\}$. The new measurement, $a<b$, does not contradict either the first or third partition. Thus the first and third partitions are already valid samples from $P_{n+1}$ for the front and back of the list. If the sample fails the test, then we need only resample the middle partition $\{b...a\}$. This middle partition was sampled from $P_n$, and consequently is already roughly in order of decreasing $n_{dispute}$. Reversing the sublist, the elements will be in an efficient order for our maximum element sampler.

If we choose a comparison at random to ask the user, then we would expect the length of the middle partition $d \sim L/3$. Therefore naively, one might expect the scaling of generating samples to remain $O(L^2)$. However we do not ask the user to make randomly chosen comparisons, we ask the user to make the comparison with the maximum expected information content. Intuitively, it is easy for the algorithm to learn the order of two elements if they lie far apart in the user's true list. However it is harder for the algorithm to learn the order of two elements which lie close together in the true list. Consequently, as the number of measurements $n$ increases, we expect the typical length $d$ of the middle partition to fall; and so the time required to generate a sample may fall below $O(L^2)$ as the number of measurements $n$ increases.

\section{'Discrete Adiabatic' Monte Carlo}
Our complete algorithm proceeds as follows,
\begin{itemize}
\item Generate $N$ random candidate lists from the uniform distribution.
\item Identify the two elements $\{ij\}$ about which the candidate orderings are most uncertain, such that we minimise $( N_{ij} - N_{ji})^2$, where $N_{ij}$ is the number of candidate lists with $i<j$.
\item Ask the user if $i <j$? We assume without loss of generality that the response is "yes, $i < j$".
\item For each candidate, check if it is is consistent with the measurement. If it is, keep it. If it is not, keep it with probability $(1-p/p)$. If the candidate was rejected, it must have structure $\{a...bj...ic...d\}$. Keep the first and final sublists $\{a...b\}$ and $\{c...d\}$, and resample the middle section $\{j...i\}$ using our maximum element sampler\footnote{When the number of measurements $n$ is small, it will be more efficient to use the recursive sampler.}.
\item Return to step 2, and terminate the algorithm when all $N$ candidate orderings are the same list.
\end{itemize}

This is a 'discrete adiabatic' algorithm, in the sense that our Monte Carlo samples are not redrawn between each measurement, but obtained by modifying the preceding samples. Each measurement conveys a small amount of information, and the probability distribution of the samples changes gradually as the number of measurements increases. Note that, for any finite $N$, we cannot guarantee that the algorithm will converge on the true list, but that as $N\rightarrow \infty$, the probability that the algorithm converges on an incorrect list, $P_{failure} \rightarrow 0$.

New samples must be redrawn between each measurement. We have shown that this requires at most $O(N L^2)$ computational time, and potentially less. Additionally we must also select the best two elements $\{ij\}$ for the user to compare. Since there are $L(L+1)/2$ possible pairs of elements, this can also be performed in $O(N L^2)$ time. We could reduce this scaling, by selecting $\{ij\}$ from a fraction of the possible pairings.  We show in the appendix that we could plausibly explore $O(L)$ of the $O(L^2)$ possible pairs without significantly impeding the overall performance of the algorithm. 

We also show in the appendix that, as expected, the algorithm requires $O(L \ln L)$ comparisons to identify the true list. Consequently, the overall computational scaling is at most $O(NL^3 \ln L)$\footnote{When N is large we do not terminate when all candidates are identical, but when the fraction of identical samples $f > 1- \epsilon$, where $\epsilon$ controls the tolerable failure rate and $0 < \epsilon \ll 1$.}, and potentially lower\footnote{For the reasons stated in the main text and appendix, we believe that the overall computational scaling can be reduced to $O(NL^2 \ln L)$ without impeding performance. Alternatively, this could be achieved by taking $O(L)$ measurements between each update of the sample lists. Furthermore, it is likely that an $O(N L \ln L)$ algorithm can be devised with only a modest increase in the number of human judgements required. However to proceed, we would need to quantify the effect \textit{"selection from previous candidates"} has on the computational cost of updating the samples between measurements.}. This sounds rather disappointing, but our goal was not to minimise the overall computational scaling; it was to minimise the number of comparisons required. We chose this goal because the cost of collecting a single human judgement significantly exceeds the cost of a single computational step. The overall cost of the algorithm is at most,
\begin{equation}
C(N, L) \leq \alpha L \ln L + \beta NL^3 \ln L
\end{equation}
Where $\alpha$ denotes the proportionality constant of human judgements, $\beta$ denotes the proportionality constant of computational calculations, and $\alpha \gg \beta$. It will remain logical to minimise the number of human judgements required so long as $L < (\alpha/\beta N)^{1/2}$. The larger $N$, the more likely we are to chose optimal comparisons; consequently $\alpha$ is a weakly declining function of $N$. In practice we expect $N \gtrsim 100$.

\section{Handling an unknown error rate}
Thus far, we have assumed that the probability of an accurate measurement, $p$, is known. This will often be unrealistic. Bayes theorem tells us that,
\begin{eqnarray}
P_n(\text{order,}p | \text{data}) &=& P_n(\text{data}|\text{order,}p) P(p) / Z \\
 &=& p^{n_{match}} (1-p)^{n-n_{match}}/Z \nonumber
\end{eqnarray} 
$P(p)$ is the prior probability distribution for the error rate, and we assume in the second equation that $P(p) = 1$ for all $0 \leq p \leq 1$; such that we begin with no idea how reliable the measurements are. However we are not directly interested in inferring $p$, but only in inferring the ordering of the true list,
\begin{eqnarray}
P_n(\text{order} |\text{data}) &=& \int_0^1 dp \, P_n(\text{order,}p | \text{data}) \\
&\propto& \int_0^1dp \,p^{n_{match}} (1-p)^{n-n_{match}}  \nonumber \\
&\propto& (1+n_{match})! (1+ n - n_{match})!   \nonumber
\end{eqnarray}
We have neglected the denominator $(2+n)!$ in the final equation, since this is independent of the order of the elements. The crucial property which underlay our previous algorithm was that,
\begin{eqnarray}
P_{n+1} \propto  P_n  \times f_n &&\text{if candidate matches measurement,} \nonumber \\
P_{n+1} \propto  P_n \times (1-f_n) &&\text{otherwise}.
\end{eqnarray}
Where $0 \leq f_n \leq 1$. In the previous case, $f_n = p$ was constant. In this case $f_n$ is not constant but it can be evaluated at each step,
\begin{equation}
f_n = \frac{2+n_{match}}{4 + n}.
\end{equation}
Thus it is possible to extend our earlier algorithm to sort a list based on human judgements, even if we have no idea how reliable the judgements are, so long as the error rate is constant and independent of the elements being compared. Some care must be taken to keep track of $n$ and $n_{match}$. If a sample is generated from scratch, then both $n$ and $n_{match}$ must be set to zero, and then gradually incremented as the sample is generated by passing a series of "tests" one by one.

\section{Conclusions}
There are many tasks which cannot be automated, since they depend on the skilled judgement or subjective opinion of a human customer or expert. However people suffer from many cognitive biases, and are extremely poor at performing reliable inferences. Computers are extremely good at performing inferences, so long as a reliable probability model exists for the problem at hand. Taking the classic problem of sorting lists, we demonstrate that an efficient algorithm can be designed which interfaces between human judgements and a computational inference engine. In this work we considered a simple probability model, however we hope to consider more complex probability models in future work. This problem is particularly elegant, in that it combines foundational ideas across computer science, information theory and inference. There is great commercial interest within the tech community at present in training employees with skills which bridge this divide, and we hope that this work may have some pedagogical value.

\acknowledgements

I have benefited from many helpful conversations in the course of this work, and would like to thank Pascal Bugnion, Marc Warner, Will Handley, John Sturm, Andrew Morris, Victor Jouffrey, Daniel Rowlands, David Turban, Robert Baldock and John Biggins for their insights. I would also like to thank my PhD supervisor Alex Chin, and acknowledge funding from the Winton Programme for the Physics of Sustainability. Finally, this work would not have been possible without David Mackay's excellent and freely available online lectures on "Information Theory, Pattern Recognition and Neural Networks".

\appendix*

\section{}
We will now demonstrate that the list can be sorted using $O(L \ln L)$ measurements, even when each measurement has a finite probability of error. 

There are $L!$ possible orderings of the list. Each order defines $L(L-1)/2$ pairwise relationships. For example, $\{abc\}$ implies $a<b$, $a<c$ and $b<c$. Only one ordering is correct. For this ordering, all of its associated pairwise relationships are also correct. The remaining $(L! - 1)$ lists are all incorrect. Some of their associated pairwise relationships are correct, but at least one pairwise statement in each incorrect ordering is false. We can define a probability distribution $P(f)$, $0 \leq f \leq 1$, which defines the probability that a fraction $f$ of the pairwise relationships defined by a randomly chosen ordering are actually true,

We do not know this probability distribution, but we know that it is symmetric about $f_{mode} = 1/2$, since every list can be read backwards. We also know that there is only one ordering with no errors (the true list). Crucially, as $L \rightarrow \infty$, we expect the distribution $P(f)$ to be increasingly sharply peaked about $f_{mode} = 1/2$. Therefore as $L \rightarrow \infty$, an overwhelmingly high fraction of the $L!$ possible orderings become clustered in a vanishingly small region, such that the fraction of true pairwise statements, $f = 1/2 \pm \delta$. This defines the typical set of incorrect orderings, for which we can approximate $f_{typical} = 0.5$. Regardless of the error rate $p$, we therefore expect $n_{match} \sim n/2$ for each of the $O(L!)$ orderings in the typical set, \textit{if we chose our measurements randomly}. Meanwhile, for the true list, $f_{true} = 1$, and we expect $n_{match} \sim np$. The probability that we will sample the true list, and not a list from the typical set,
\begin{eqnarray}
P_n(\text{true}|\text{data}) &\sim& \frac{P_n(\text{data}|\text{true})}{P_n(\text{data}|\text{true}) + L! \times P_n(\text{data}|\text{typical})} \nonumber \\
       &=& \frac{p^{np}(1-p)^{n(1-p)}}{p^{np}(1-p)^{n(1-p)} + L! \times p^{n/2}(1-p)^{n/2}} \nonumber \\
       &=& \left(1 + L! \times (p/(1-p))^{-n(p-1/2)} \right)^{-1}. \nonumber
\end{eqnarray}
We have successfully eliminated the typical set when $P_n(\text{true}|\text{data}) > 1- \epsilon$, where $\epsilon$ is a small constant defining the tolerable failure rate. Using the binomial expansion, we expect to reach this failure rate once,
\begin{eqnarray}
L! \times (p/(1-p))^{-n(p-1/2)} \sim \epsilon. \nonumber
\end{eqnarray}
We are interested in the number of measurements $n$ required to reach this point. Rearranging and applying Sterling's formula, $n \sim O(L \ln L - \ln \epsilon)$. It is indeed possible to exclude the typical set of incorrect alternative orderings with $O(L \ln L)$ measurements.

However, we have not yet proved that the list can be sorted correctly in $O(L \ln L)$ measurements. Although we have excluded the typical set, we have not excluded the atypical set. This atypical set is composed of the vanishingly small fraction of incorrect orderings for which $f \sim 1$. Although there are far fewer of these orderings, they are ultimately the most difficult to exclude, since their predictions are highly consistent with the measurements observed. For simplicity, we will only consider the most difficult incorrect orderings of all; the single flip errors. These can be found by taking the true ordering, and interchanging two neighbouring elements. There are $(L-1)$ such orderings. All but one of the pairwise statements associated with these lists are true. Thus as $L \rightarrow \infty$, $f_{atypical} \rightarrow 1 - 2/L^2$. If we continue to make randomly chosen measurements,
\begin{eqnarray}
&&P_n(\text{true}|\text{data}) \sim \frac{P_n(\text{data}|\text{true})}{P_n(\text{data}|\text{true}) + L \times P_n(\text{data}|\text{atypical})} \nonumber \\
 &=& \frac{p^{np}(1-p)^{n(1-p)}}{p^{np}(1-p)^{n(1-p)} + L p^{n(p-\frac{4p-2}{L^2})}(1-p)^{n(1-p+\frac{4p-2}{L^2})}} \nonumber \\
       &=& \left(1 + L  (p/(1-p))^{-4n(p-1/2)/L^2} \right)^{-1}. \nonumber
\end{eqnarray}
We expect our algorithm to terminate once $L (p/(1-p))^{-4n(p-1/2)/L^2} \sim \epsilon$. Rearranging, $n \sim O(L^2 (\ln L - \ln \epsilon))$. Thus it appears that we require $O(L^2 \ln L)$ measurements to exclude the atypical set of single flip errors. However this is a consequence of making randomly chosen measurements. There are $O(L^2)$ possible measurements, but only $O(L)$ of them enable us to distinguish the true list from the atypical set. In the case of single flip errors, these $O(L)$ useful measurements correspond to the $L-1$ comparisons between neighbouring elements in the true list. If we do not choose our measurements at random, but instead use our "greedy entropy" algorithm to select useful comparisons, then we will reduce the number of measurements required by $O(L)$, leading to an overall scaling of $O(L \ln L)$ as desired.

We noted in the main text that since there are $O(L^2)$ possible comparisons, it takes $O(NL^2)$ computational time to evaluate which measurement should be made next; implying that the overall computational cost of the algorithm is at least $O(NL^3 \ln L)$, regardless of how efficiently we can generate samples from $P_n(\text{order}|\text{data})$. However we now see that we can reduce this cost to $O(NL)$, by only considering pairs of elements $\{ij\}$ which are adjacent in one of the current candidate orderings. This will not significantly impede the performance of the algorithm, since random measurements are sufficient to exclude the typical set, while the relevant measurements to exclude the atypical set correspond to neighbouring elements in the true list. After the typical set is excluded, the candidate orderings will increasingly resemble the unknown true list, and this procedure will successfully identify the most informative measurements.

\bibliography{Sorting}

\begin{thebibliography}{6}
\expandafter\ifx\csname natexlab\endcsname\relax\def\natexlab#1{#1}\fi
\expandafter\ifx\csname bibnamefont\endcsname\relax
  \def\bibnamefont#1{#1}\fi
\expandafter\ifx\csname bibfnamefont\endcsname\relax
  \def\bibfnamefont#1{#1}\fi
\expandafter\ifx\csname citenamefont\endcsname\relax
  \def\citenamefont#1{#1}\fi
\expandafter\ifx\csname url\endcsname\relax
  \def\url#1{\texttt{#1}}\fi
\expandafter\ifx\csname urlprefix\endcsname\relax\def\urlprefix{URL }\fi
\providecommand{\bibinfo}[2]{#2}
\providecommand{\eprint}[2][]{\url{#2}}

\bibitem[{\citenamefont{Feige et~al.}(1990)\citenamefont{Feige, Peleg,
  Raghavan, and Upfal}}]{feige1990computing}
\bibinfo{author}{\bibfnamefont{U.}~\bibnamefont{Feige}},
  \bibinfo{author}{\bibfnamefont{D.}~\bibnamefont{Peleg}},
  \bibinfo{author}{\bibfnamefont{P.}~\bibnamefont{Raghavan}}, \bibnamefont{and}
  \bibinfo{author}{\bibfnamefont{E.}~\bibnamefont{Upfal}}, in
  \emph{\bibinfo{booktitle}{Proceedings of the twenty-second annual ACM
  symposium on Theory of computing}} (\bibinfo{organization}{ACM},
  \bibinfo{year}{1990}), pp. \bibinfo{pages}{128--137}.

\bibitem[{\citenamefont{Busse et~al.}(2012)\citenamefont{Busse, Chehreghani,
  and Buhmann}}]{busse2012information}
\bibinfo{author}{\bibfnamefont{L.~M.} \bibnamefont{Busse}},
  \bibinfo{author}{\bibfnamefont{M.~H.} \bibnamefont{Chehreghani}},
  \bibnamefont{and} \bibinfo{author}{\bibfnamefont{J.~M.}
  \bibnamefont{Buhmann}}, in \emph{\bibinfo{booktitle}{Information Theory
  Proceedings (ISIT), 2012 IEEE International Symposium on}}
  (\bibinfo{organization}{IEEE}, \bibinfo{year}{2012}), pp.
  \bibinfo{pages}{2746--2750}.

\bibitem[{\citenamefont{Alonso et~al.}(2004)\citenamefont{Alonso, Chassaing,
  Gillet, Janson, Reingold, and Schott}}]{CPC:248097}
\bibinfo{author}{\bibfnamefont{L.}~\bibnamefont{Alonso}},
  \bibinfo{author}{\bibfnamefont{P.}~\bibnamefont{Chassaing}},
  \bibinfo{author}{\bibfnamefont{F.}~\bibnamefont{Gillet}},
  \bibinfo{author}{\bibfnamefont{S.}~\bibnamefont{Janson}},
  \bibinfo{author}{\bibfnamefont{E.~M.} \bibnamefont{Reingold}},
  \bibnamefont{and} \bibinfo{author}{\bibfnamefont{R.}~\bibnamefont{Schott}},
  \bibinfo{journal}{Combinatorics, Probability and Computing}
  \textbf{\bibinfo{volume}{13}}, \bibinfo{pages}{419} (\bibinfo{year}{2004}).

\bibitem[{\citenamefont{Ajtai et~al.}(2016)\citenamefont{Ajtai, Feldman,
  Hassidim, and Nelson}}]{ajtai2009sorting}
\bibinfo{author}{\bibfnamefont{M.}~\bibnamefont{Ajtai}},
  \bibinfo{author}{\bibfnamefont{V.}~\bibnamefont{Feldman}},
  \bibinfo{author}{\bibfnamefont{A.}~\bibnamefont{Hassidim}}, \bibnamefont{and}
  \bibinfo{author}{\bibfnamefont{J.}~\bibnamefont{Nelson}},
  \bibinfo{journal}{ACM Transactions on Algorithms (TALG)}
  \textbf{\bibinfo{volume}{12}}, \bibinfo{pages}{19} (\bibinfo{year}{2016}).

\bibitem[{\citenamefont{Hadjicostas and
  Lakshmanan}(2011)}]{hadjicostas2011recursive}
\bibinfo{author}{\bibfnamefont{P.}~\bibnamefont{Hadjicostas}} \bibnamefont{and}
  \bibinfo{author}{\bibfnamefont{K.}~\bibnamefont{Lakshmanan}},
  \bibinfo{journal}{Discrete Applied Mathematics}
  \textbf{\bibinfo{volume}{159}}, \bibinfo{pages}{1398} (\bibinfo{year}{2011}).

\bibitem[{\citenamefont{Yao and Yao}(1985)}]{yao1985fault}
\bibinfo{author}{\bibfnamefont{A.~C.} \bibnamefont{Yao}} \bibnamefont{and}
  \bibinfo{author}{\bibfnamefont{F.~F.} \bibnamefont{Yao}},
  \bibinfo{journal}{SIAM Journal on Computing} \textbf{\bibinfo{volume}{14}},
  \bibinfo{pages}{120} (\bibinfo{year}{1985}).

\end{thebibliography}

\end{document}